\documentclass[manuscript]{acmart}

\AtBeginDocument{%
  }

\setcopyright{acmlicensed}
\copyrightyear{2025}
\acmYear{2025}
\acmDOI{XXXXXXX.XXXXXXX}

\acmConference[Conference acronym 'XX]{Make sure to enter the correct
  conference title from your rights confirmation emai}{June 03--05,
  2018}{Woodstock, NY}
\acmISBN{978-1-4503-XXXX-X/25/06}

\begin{document}
\title{From \#Dr00gtiktok to \#harmreduction: Exploring Substance Use Hashtags on TikTok}

\author{Layla Bouzoubaa}
\authornote{Equal contribution}
\email{lb3338@drexel.edu}

\author{Muqi Guo}
\authornotemark[1]

\author{Joseph Trybala}

\author{Afsaneh Razi}

\author{Rezvaneh Rezapour}

\affiliation{%
  \institution{Department of Information Science, Drexel University}
  \city{Philadelphia}
  \state{Pennsylvania}
  \country{USA}
}
  
\label{sec:abstract}
\begin{abstract}
TikTok has emerged as a major source of information and social interaction for youth, raising urgent questions about how substance use discourse manifests and circulates on the platform. This paper presents the first comprehensive analysis of publicly visible, algorithmically surfaced substance-related content on TikTok, drawing on hashtags spanning all major substance categories. Using a mixed-methods approach that combines social network analysis with qualitative content coding, we examined 2,333 substance-related hashtags, identifying 16 distinct hashtag communities and characterizing their structural and thematic relationships.
Our network analysis reveals a highly interconnected small-world structure in which recovery-focused hashtags such as \textit{\#addiction}, \textit{\#recovery}, and \textit{\#sober} serve as central bridges between communities. Qualitative analysis of 351 representative videos shows that Recovery Advocacy content (33.9\%) and Satirical content (28.2\%) dominate, while direct substance depiction appears in only 26\% of videos, with active use shown in just 6.5\% of them. These findings suggest that the algorithmically surfaced layer of substance-related discourse on TikTok is predominantly oriented toward recovery, support, and coping rather than explicit promotion of substance use.
We further show that hashtag communities and video content are closely aligned, indicating that substance-related discourse on TikTok is shaped through organic community formation within platform affordances rather than widespread adversarial evasion of moderation. This work contributes to social computing research by showing how algorithmic visibility on TikTok shapes the organization of substance-related discourse and the formation of recovery and support communities.
\end{abstract}


\maketitle

\section{Introduction}
\label{sec:introduction}

Social media platforms have become critical and contested spaces for health discourse, particularly around stigmatized topics like substance use \cite{russell_using_2021,johnsonContentAnalysisPortrayal2025}. TikTok's algorithmic recommendation system has demonstrated serious harms: amplifying health misinformation \cite{yeungTikTokAttentionDeficitHyperactivity2022}, promoting eating disorders \cite{blackburnForYouImpactProana2024}, and contributing to mental health crises among young users \cite{conteScrollingAdolescenceSystematic2025}. Yet these same algorithmic affordances enable the formation of peer support communities around substance use recovery, harm reduction, and lived experience sharing \cite{russell_using_2021, goyalComprehensiveCrossSectionalAnalysis2025}. This tension reflects broader dynamics of platform governance, in which algorithmic visibility simultaneously mediates risk, care, and community formation in stigmatized health contexts, yet remains underexplored for substance use discourse, where both risks and potential benefits are acute.

\begin{figure}[h]
\centering
\begin{minipage}[t]{0.45\columnwidth}
    \centering
    \includegraphics[width=\columnwidth]{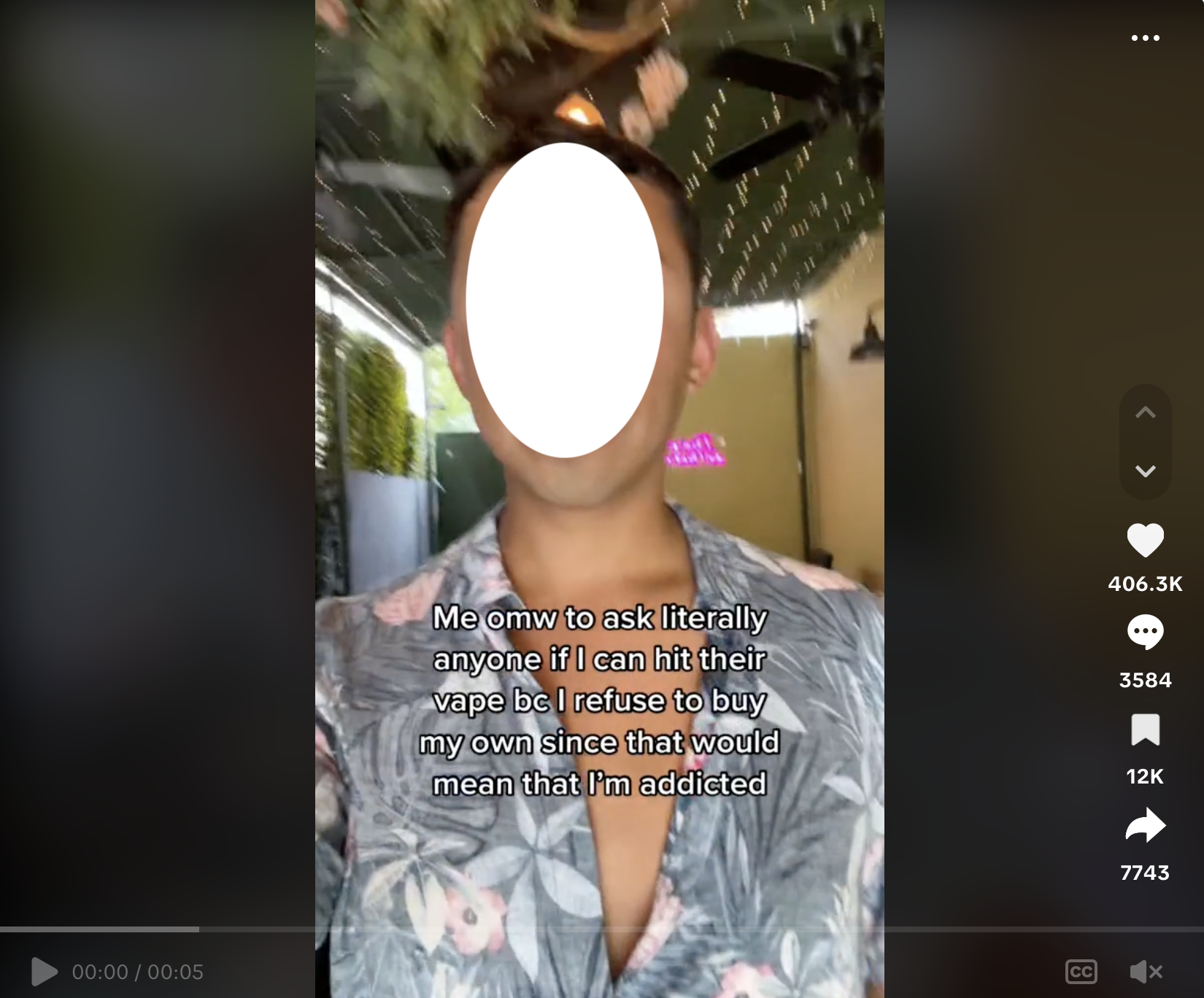}
\end{minipage}
\hspace{0.05\columnwidth}
\begin{minipage}[t]{0.45\columnwidth}
    \centering
    \includegraphics[width=\columnwidth]{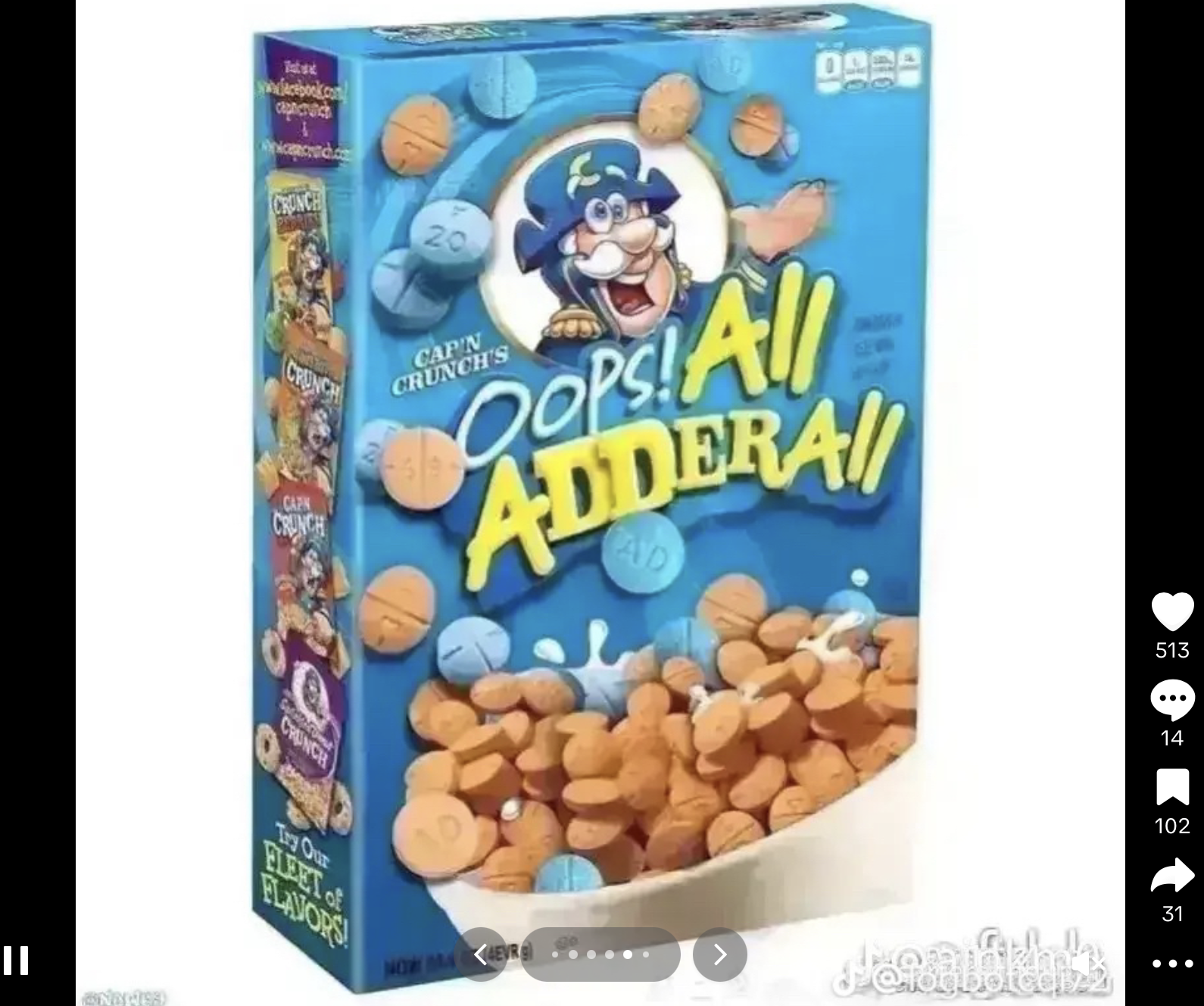}
\end{minipage}
\caption{Examples of TikTok videos illustrating distinct substance-related content identified through hashtags.}
\label{fig:tiktok_examples}
\vspace{-0.5cm}
\end{figure}

Understanding how substance use content manifests on platforms like TikTok is critical for multiple stakeholders. Early exposure to substance use significantly increases the risk of developing substance use disorders (SUD) \cite{mccabe2022longitudinal, volkow2022association, odgers2008important}, making youth exposure a public health concern. However, platforms also serve people already living with SUD who may lack offline support systems \cite{russell_using_2021, Bouzoubaa_Young_Rezapour_2024, goyalComprehensiveCrossSectionalAnalysis2025}. Previous trends like the ``xandemic'' (2016-2018), characterized by high school students documenting the recreational use of Xanax to the point of blackout, and \textit{\#pingtok}, a portmanteau of the slang term for an MDMA pill, \textit{pinger}, and TikTok, demonstrate how substance-related content can become embedded in platform culture \cite{fuller2024understanding, rutherford_changes_2022}. What remains unclear is how substance normalization and recovery-support dynamics coexist, intersect, and vary across substances and communities. 

Previous research on substance-related content on TikTok has typically focused on either specific substances such as cannabis or nicotine \cite{rutherford_turnttrending_2023, rutherford_changes_2022, morales_nicotineaddictioncheck_2022, tanPuffBarHowTop2021} or isolated trending phenomena such as \textit{\#pingtok} \cite{Whelan_Noller_Ward_2024}. While some researchers have examined hashtag networks and content moderation strategies around other sensitive topics like eating disorders \cite{Bickham2024HiddenInPlainSight} and how hashtags create communities around shared experiences \cite{eriksson_krutrok_algorithmic_2021}, there is limited comprehensive knowledge of how substance-related discourse is organized across TikTok's hashtag-driven ecosystem. In contrast to prior work that examines individual substances, isolated trends, or moderation outcomes, we take a platform-wide, cross-substance perspective that captures how substance discourse is structurally organized and socially enacted across TikTok’s hashtag ecosystem. To address this gap, we draw on an established drug classification system \cite{DEA_Factsheet} in order to examine content from every major substance category on TikTok. Our research questions are:

\begin{itemize}

    \item RQ1: What is the network structure of substance-related hashtags on TikTok, and what distinct communities emerge within this network? 
    \item RQ2: What are the topics and key characteristics of the videos associated with the hashtags in these communities? 
    \item RQ3: How does substance visibility, operationalized through engagement and algorithmic surfacing, affect topic distribution and engagement metrics?
\end{itemize}

We address these questions through a mixed-methods analysis combining social network 
analysis of hashtag co-occurrence patterns with qualitative content coding. Analyzing 2,333 exemplar hashtags and coding 351 representative videos, we examine both the structural organization of substance-related discourse and the actual content users create. Our network analysis reveals a highly interconnected discourse where recovery-focused hashtags like \textit{\#addiction}, \textit{\#recovery}, and \textit{\#harmreduction} serve as bridges between thematic communities. Qualitative coding shows that Recovery Advocacy (33.9\%) and Humor-based coping (28.2\%) make up the majority of videos, while direct substance depiction appears in only 26\% of videos, with active use shown in just 6.5\%. Our findings suggest that substance-related content on TikTok, within the publicly visible, algorithmically surfaced layer accessible to observation, is predominantly oriented toward support rather than promotion or glamorization.

Through this work, we contribute to social computing and substance use literature by providing the first comprehensive mapping of substance-related content on TikTok across all major drug categories, revealing 16 distinct hashtag communities within a highly interconnected small-world network. Second, we demonstrate that hashtag communities exhibit organic thematic coherence rather than adversarial evasion of content moderation, suggesting these support communities form within rather than against platform affordances. Third, we show that recovery advocacy (33.9\%), humor-based coping (28.2\%), and educational information (13.1\%) collectively make up most of the content, with active substance use depiction making up only 6.5\% in algorithmically surfaced public discourse. Together, these contributions advance social computing research on algorithmic visibility, health discourse, and community formation under platform governance constraints. Finally, we derive design implications for content moderation strategies that must balance harm prevention with preservation of beneficial peer support communities. These findings inform platform governance approaches, the development of social media-based recovery interventions, and youth substance use prevention efforts that account for how substance-related content actually manifests on TikTok.
\section{Related Work}
\label{sec:background}


As digital platforms become ubiquitous, understanding their impact, especially on young people's health behaviors regarding substance use, is more crucial than ever. 
On the one hand, social media emerged as a medium for facilitating sensitive discussions and disclosure of lived experiences, including mental health and eating disorders \cite{rutherford_changes_2022, Bickham2024HiddenInPlainSight, bouzoubaa-etal-2024-decoding}. On the other hand, several studies have examined the relationship between online peer groups, social media homophily, and substance use, finding positive correlations between substance use and social media engagement. For example, \citet{willoughby_exposure_2024} found that young adults reported an increased intention to use cannabis after exposure to pro-cannabis messages online. Similarly, \citet{sun_longitudinal_2023} showed that exposure to e-cigarette advertisements on social media platforms is associated with adolescents' e-cigarette usage in later years. 
\citet{miller_online_2021} examined how online and offline peer interactions influence substance use behaviors and found that online belonging and social media homophily predicted regular stimulant and opioid use among substance users. \citet{hanson_tweaking_2013} have studied online discussions related to stimulant drug use on Twitter and found a high correlation between the popularity of such topics and final exam periods. \citet{geusens_first_2023} found that higher levels of alcohol and marijuana consumption predicted later social media posts about those substances, but posting did not predict later substance use. \citet{sun_longitudinal_2023} found a positive correlation between early exposure to e-cigarette advertisement and subsequent lifetime usage. 


Hashtags, a medium for interactions on platforms like Facebook or Twitter, have been studied comprehensively in the past~\cite{mousavi_detecting_2021, ince_social_2017, small_what_2011,rezapour2017identifying}. At their core, hashtags facilitate content discovery and community formation by enabling users to collectively construct and navigate shared narratives through keyword tagging \cite{Yang_2016, small_what_2011}. 
Previous studies have examined substance use discourse across various social media platforms, contributing to our understanding of how substance use behaviors manifest online. For instance, on Instagram, investigating \textit{\#studydrugs} demonstrated a predominantly positive portrayal of stimulant use for academic performance enhancement \cite{petersen_studydrugspersuasive_2021}.

On TikTok, content creators deliberately employ hashtags like \textit{\#fyp} (For You Page) to increase visibility and maximize audience reach within TikTok's recommendation system \cite{abidin_mapping_2020}. The platform's ranking algorithms connect users based on hashtag engagement patterns, creating what \citet{eriksson_krutrok_algorithmic_2021} calls ``algorithmic closeness'' - subcommunities united through shared content interests. This algorithmic dynamic shapes how substance-related discourse spreads and evolves on TikTok. For example, research examining TikTok hashtags like \textit{\#puffbar} has revealed concerning patterns of youth-oriented substance marketing and normalization \cite{tanPuffBarHowTop2021}, while analysis of \textit{\#stonertok} and \textit{\#420vibes} showed the majority of cannabis-related videos positively portray substances \cite{rutherford_getting_2022}.

Our study advances existing research by employing a network-based and qualitative approach to explore the interactions and communities associated with substance-related hashtags on TikTok. While prior research has examined individual substances or discrete hashtag communities, the structural relationships and interactions among these communities within TikTok's broader ecosystem remain underexplored. Rather than focusing on message content or sentiment alone, we examine how substance-related discourse is structurally organized and interconnected at the platform level.
Our work provides empirical grounding for understanding algorithmic visibility, community formation, and the coexistence of harm and support within substance-related discourse, informing not only content moderation but also platform governance, intervention design, and the development of evidence-based recommendations for public health and recovery-oriented stakeholders.
\section{Methods}
\label{sec:methods}

\begin{figure}[t]
    \centering
    \includegraphics[width=0.75\linewidth]{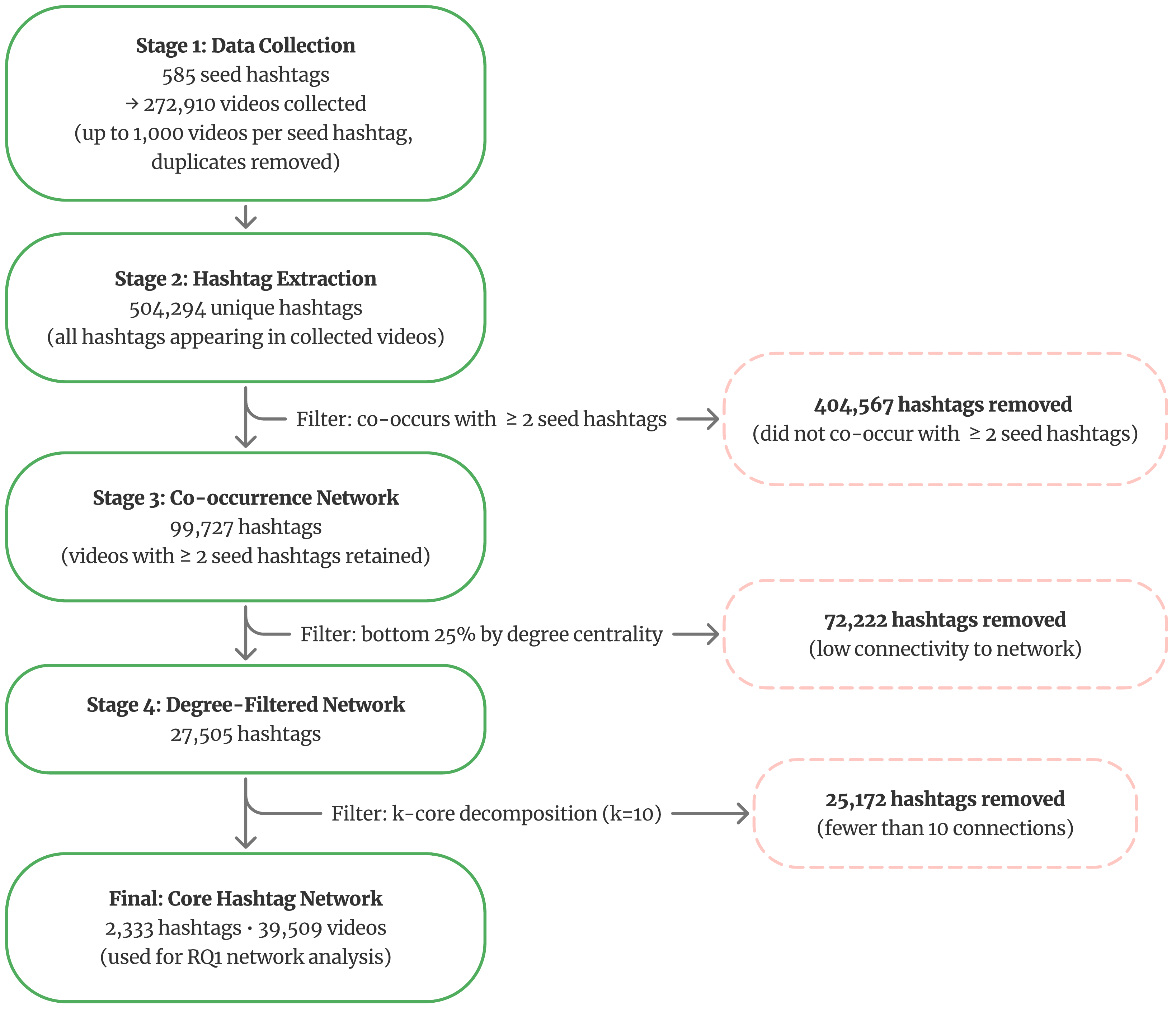}
    \caption{Data selection and pre-processing pipeline to create the filtered set.}
    \label{fig:data_selection}
    \vspace{-0.45cm}
\end{figure}

\subsection{Data}
\subsubsection{Seed Hashtag Selection: }To develop an initial set of seed hashtags related to substance use for the data collection, we first identified hashtags previously used in studies of substance-related content on social media platforms~\cite{petersen_studydrugspersuasive_2021, russell_analyzing_2022, rutherford_turnttrending_2023, rutherford_getting_2022, singh_sentiment_2021} as well as hashtags associated with substance names (including both formal and slang terms) listed on the Drug Enforcement Agency's Fact Sheet \cite{DEA_Factsheet}. 
Using this set, we conducted iterative searches on TikTok to identify additional relevant hashtags by examining co-occurring hashtags within the content. This process continued until we reached saturation (no substantively new substance-related hashtags yielded), resulting in a final set of 585 substance-use-related hashtags.\footnote{The list will be shared upon the paper's acceptance.}

\subsubsection{Video Collection: }Using the Unofficial TikTok API\footnote{\url{https://github.com/davidteather/TikTok-Api}}, we collected  up to 1,000 publicly available, algorithmically surfaced videos for each of the 585 seed hashtags during the data collection period.
This resulted in an initial, raw dataset of 272,910 videos. These videos contained a total of 504,294 unique hashtags (including both substance-related terms and general hashtags that co-occurred with them).

\subsubsection{Network Construction and Refinement:} To enhance the feasibility of our analysis and reduce noise from potentially irrelevant content, we filtered the data to include only videos that featured at least two of the seed hashtags, ensuring that retained videos reflected explicit engagement with substance-related discourse rather than incidental hashtag use. This filtering step resulted in a refined dataset of 99,727 unique hashtags (Figure \ref{fig:data_selection}).
To further clean the data, we constructed a hashtag co-occurrence network where nodes represent individual hashtags and edges represent their co-occurrence in the same video. Edge weights were defined as the frequency with which two hashtags appeared together in videos. 

We then employed a two-stage network refinement process to focus on the most relevant and interconnected hashtags.
First, we removed hashtags with low connectivity by filtering out nodes in the bottom 25th percentile of degree centrality \cite{borgatti2005centrality}. This step eliminated peripheral hashtags with minimal connections to the broader substance use discourse, removing those that were weakly connected or contextually incidental rather than substantively related to substance use discussions. 
Next, we applied k-core decomposition ($k=10$) to identify the densely connected core of the network \cite{seidman1978graph}. K-core decomposition recursively removes nodes with degrees less than $k$ until all remaining nodes have at least $k$ connections, ensuring a cohesive network structure. The value of $k=10$ was selected based on exploratory analysis, which indicated that this threshold preserved network connectivity while excluding sparsely connected hashtags. The resulting refined network of hashtags contained 2,333 nodes and 46,990 edges, representing the most interconnected hashtags in substance use-related content. 

\begin{table*}[t]
\centering
\scriptsize 
\begin{tabular}{|p{2.7cm}|p{8cm}|p{2.7cm}|}
\hline
\textbf{Communities} & \textbf{Definition} & \textbf{Example Hashtags} \\ \hline
Emotions and Feelings & Words and phrases related to emotional states and feelings. & \#love, \#gratitude \\ \hline
Health Conditions & Words and phrases related to health issues directly related to substance use, addiction-related health problems, or chronic conditions that may lead to substance use. 
& \#addiction, \#chronicpain \\ \hline
Alcohol & Words related to alcoholic beverages and spirits. This is just about the substances or drinks and not the side effects and consumption methods. & \#alcohol, \#vodka \\ \hline
Cannabis & Substances related to marijuana, including recreational and medicinal use. & \#cannabis, \#stonersoftiktok \\ \hline
Cognitive Enhancement & Substances about nootropics, smart drugs, and methods to improve cognitive function. & \#modafinil, \#smartdrugs \\ \hline
Commonly-Misused Substances & Substances frequently used and misused, such as licit and illicit substances. & \#heroin, \#xans \\ \hline
Consumption Method & Words and phrases that show specific ways in which substances (including alcohol, illicit or licit drugs, tobacco, and cannabis) are ingested, administered, or used. & \#smoking, \#injection \\ \hline
Awareness and Advocacy & Words related to information and strategies to prevent and raise awareness of substance abuse or reduce associated harm, discussions about drug laws, policies, and related social issues, and content about addiction recovery, sobriety, and support systems. & \#harmreduction, \#recoverylife \\ \hline
Other Substances & Mentions of less common substances, oftentimes legal, over-the-counter medications, herbal remedies, or supplements. & \#vitamins, \#magnesium \\ \hline
Platform & Tags and features specific to social media engagement, visibility, and trending tactics. & \#fyp, \#viral \\ \hline
Substance Effects & Words and phrases that describe the physical or mental effects of substance or alcohol use, both desired effects and side effects. & \#stoned, \#drunk \\ \hline
Tobacco Nicotine & Words related to tobacco products, cigarettes, vaping, and nicotine use. & \#vaping, \#ecigs \\ \hline
Humor & Words, phrases, or hashtags related to jokes, memes, or any content meant to be funny but specific to substance use, addiction, or recovery. & \#soberhumor, \#drughumor \\ \hline
Location & Words related to geographical locations, including cities, states, countries, or continents. & \#kensingtonphilly, \#boston \\ \hline
Occupation & Words related to occupations or professions. & \#nurselife, \#medstudent \\ \hline
Identity and Community & Hashtags related to any social identity, demographic group, or community affiliation, such as race, ethnicity, gender identity, sexual orientation, disability status, socioeconomic background, immigration status, religion, age group, or membership in specific subcultures or communities. & \#lgbtqia, \#transgender \\ \hline
Misc & Any tag that does not fit into the above categories. & \#cookies, \#foundersday \\ \hline
\end{tabular}
\caption{Categories for semantically classifying hashtags from TikTok content, their definitions, and example hashtags.}
\label{tab:themes}
\end{table*}

\subsection{Substance Use Hashtags Network Analysis (RQ1)}
\subsubsection{Community Classification:} We first applied established community detection algorithms, including Clauset-Newman-Moore \cite{Clauset2004FindingCS} and Louvain method \cite{blondel2008fast}, but these approaches did not result in clusters that meaningfully captured substance-related discussions, as the resulting communities were driven primarily by network density rather than thematic or semantic coherence. Therefore, we used a hybrid approach combining manual and automated methods to identify thematic groups within the hashtags. 
We first manually annotated a sample of 251 hashtags drawn from prior work, manual TikTok searches and DEA fact sheets \cite{DEA_Factsheet} and iteratively established substance-related content categories. This iterative process was informed by prior work on substance use discourse and refined through repeated examination of hashtag usage and contexts. As a result, we identified 16 categories representing topics related to substance use and consumption, identity, and community, health and emotion as well as awareness and advocacy. In addition, we created a Misc category for general hashtags, not related to substance use content. The final set of categories and their definitions are shown in Table \ref{tab:themes}. 
These categories were then used to prompt GPT-4o \cite{openai2024gpt4} to classify hashtags. When evaluated against human-labeled data, our best-performing prompt achieved an overall accuracy of $89.36\%$, demonstrating strong reliability in the thematic categorization. We then used this prompt to classify the full set of 2,333 hashtags. The resulting labeled hashtags were mapped back onto the network, creating themed communities that reflect different aspects of substance use discourse on TikTok.

\subsubsection{Network Analysis: }For both the overall network and each community, we calculated standard network centrality measures including betweenness \cite{freeman1977set}, closeness \cite{sabidussi1966centrality}, degree \cite{freeman1978centrality}, and eigenvector centralities \cite{bonacich1972factoring}. These widely used measures capture complementary aspects of structural prominence, including reachability, connectivity, and association with highly connected hashtags and help identify influential hashtags in terms of their structural roles within the discourse network. Additionally, we computed network-level metrics such as average degree, clustering coefficient, and average shortest path length to characterize overall connectivity, cohesion, and small-world properties of substance use–related discourse on TikTok, both at the global network level and within individual communities.

\subsection{Qualitative Analysis of Video Content (RQ2)}
Building on the hashtag communities identified in RQ1, we analyzed video content associated with the 2,333 core hashtags organized into 16 thematic clusters (excluding the Misc cluster). While network analysis reveals how substance-related discourse is structurally organized, it cannot capture the nature of content within these communities (e.g., whether videos promote use, support recovery, or serve other functions). To address RQ2, we qualitatively examined how substance-related discourse manifests within and across these communities along two dimensions: video topic and substance presence. 

\subsubsection{Codebook Development}

We developed a qualitative codebook using an iterative open-coding process designed to capture salient characteristics of substance-related video content~\cite{braun2006using}. Two members of the research team independently coded an initial set of 160 randomly selected videos drawn from the broader 39,509-video dataset. Through this exploratory open-coding process, the coders identified an initial set of nine video topic categories reflecting recurring themes in the content.
In parallel, we coded for substance presence, capturing whether substances were visually depicted or actively used within videos. This dimension was included to assess how different communities and topics varied in their explicit depiction of substances. Following this initial round, the research team met to discuss discrepancies, refine category definitions, and consolidate overlapping themes. Through another round of independent coding, reliability was assessed on a 10\% (n = 40) of the coded videos, yielding moderate agreement for video topic ($\kappa = 0.51$) and substance shown ($\kappa = 0.58$). This process resulted in a reduced set of seven video topic categories. Remaining disagreements were resolved through deliberative discussion among the full research team until consensus was reached. The 160 development videos were excluded from the final analytic sample to maintain independence between codebook development and application. The final codebook captured two primary dimensions: (1) video topic and (2) substance presence. Definitions and examples for each category are provided in Table~\ref{tab:codebook}.

\subsubsection{Video Sampling}
We constructed the qualitative sample through a multi-stage, stratified selection process designed to ensure coverage across communities and engagement levels. We started with the full set of 39,509 videos containing at least one of the 2,333 core hashtags.
To ensure representation across the 16 hashtag communities, we first selected the top five hashtags per community based on degree centrality, yielding a total of 80 focal hashtags. Degree centrality was used to prioritize hashtags that were structurally prominent within each community's discourse network. For each focal hashtag, we stratified associated videos into five engagement quintiles based on view count, calculated within each hashtag's distribution.

Our target was to code at least one video per engagement quintile per hashtag (5 videos × 80 hashtags = 400 minimum). To account for anticipated exclusions due to content removal, non-English language, or lack of substance relevance, we initially sampled up to 10 videos per quintile where available. However, many hashtags had fewer than 50 associated videos in the filtered dataset, resulting in incomplete coverage of some quintile-hashtag combinations. This process yielded 977 candidate videos for screening.
Each candidate video was then evaluated against three inclusion criteria:
\begin{itemize}
    \item \textbf{Availability}: The video was publicly accessible at the time of coding (i.e., not removed or restricted).
    \item \textbf{Language}: The primary spoken or written language of the video was English.
    \item \textbf{Substance Relevance}: The video clearly involved substance-related content, indicated through visual cues, discussion, or behaviors.
\end{itemize}

Of the 977 candidate videos, 852 (87.2\%) were publicly available at the time of analysis, 695 (81.5\% of available videos) were in English, and 351 (50.5\% of English videos) met the substance relevance criterion. These 351 videos constituted the final corpus for inductive qualitative content analysis \cite{elo2008qualitative}. The videos were divided between the two coders for independent coding. Coders consulted with the broader research team when encountering ambiguous cases to ensure consistent application of the codebook. We report descriptive frequencies for all coded dimensions. Descriptive statistics for the sampled videos are reported in Table~\ref{tab:descriptive}.

\section{Findings}
\label{sec:findings}

\begin{table}[t]
\centering
\begin{tabular}{|l|l|l|}
\hline
                       & \textbf{Filtered} & \textbf{Coded} \\ \hline
\textbf{N Videos}      & 39,509            & 351\\ \hline
\textbf{Avg. Likes}    & 66.5K            &                30.9K\\ \hline
\textbf{Avg. Comments} & 702.6             &                305.9\\ \hline
\textbf{Avg. Shares}   & 3513.2&                2652.5\\ \hline
\textbf{Min. Date}     & 2015-09-22        &                2019-11-13\\ \hline
\textbf{Max. Date}     & 2024-05-13        &                2024-05-09\\ \hline
\end{tabular}%
\caption{Video statistics of the dataset filtered for RQ1 and the sub-set used for coding in RQ2.}
\label{tab:descriptive}
\vspace{-0.5cm}
\end{table}

\subsection{RQ1: Network Structure and Communities}
Analysis of the hashtag network revealed a highly interconnected structure 
with 2,333 nodes (hashtags) and 46,990 edges (Table \ref{tab:centrality-measures} in Appendix), with an average degree of 40.28 and a weighted degree of 513.84. These metrics indicate substantial interconnection; each hashtag co-occurs with approximately 40 other hashtags, and when hashtags do co-occur, they appear together frequently (over 500 times on average). The network exhibits small-world properties \cite{watts_collective_1998}, characterized by a high average clustering coefficient (0.52) and a short average path length (2.20). This structure suggests that substance-related discussions form tightly knit clusters while remaining structurally connected through relatively short paths across the broader network.

Network centrality analysis identified recovery-focused hashtags as structurally central. The five hashtags with highest betweenness centrality: \textit{\#addiction} (0.129), \textit{\#recovery} (0.098), \textit{\#sober} (0.058), \textit{\#wedorecover} (0.050), and \textit{\#harmreduction} (0.049), all relate to recovery and harm reduction, positioning these terms as bridges connecting otherwise disparate thematic communities.

Community analysis identified 16 thematic clusters (excluding Misc, see Table \ref{tab:themes}), each reflecting distinct structural properties within substance-related discourse (Figure \ref{fig:full-graph} in Appendix). Communities varied substantially in internal cohesion.
The largest community, Awareness and Advocacy, comprises approximately 12\% of all hashtags (279 nodes) with high clustering (0.89), and includes hashtags related to harm reduction (e.g., \textit{\#harmreduction}, \textit{\#harmreductionworks}), recovery and sobriety (e.g., \textit{\#wedorecover}, \textit{\#sober}), and public health awareness (e.g., \textit{\#opioidawareness}, \textit{\#opioidcrisis}). Similarly, Health Conditions (277 nodes, clustering = 0.85) showed high internal cohesion around hashtags such as \textit{\#addiction} and \textit{\#chronicpain}.

Substance-specific communities exhibit varied structural patterns. Cannabis (51 nodes, clustering = 0.55) and Tobacco/Nicotine (31 nodes, clustering = 0.70) show high internal clustering, while Commonly-Misused Substances (152 nodes, clustering = 0.38) and Alcohol (43 nodes, clustering = 0.41) exhibit more moderate cohesion. The Commonly-Misused Substances community contains hashtags that often span awareness-oriented and substance-use-oriented content (e.g., \textit{\#studydrugs} \cite{petersen_studydrugspersuasive_2021}, \textit{\#pingertok} \cite{Whelan_Noller_Ward_2024}, \textit{\#smartdrugs} \cite{petersen_studydrugspersuasive_2021}). 

In contrast, several communities function as cross-cutting connectors with low internal clustering. The Consumption Method community (41 nodes, clustering = 0.30), containing hashtags such as \textit{\#smoke}, \textit{\#vape}, and \textit{\#injection}, exhibit the highest average betweenness centrality among all communities, connecting discussions across substance types. The Platform-Specific community (183 nodes, clustering = 0.12), characterized by hashtags such as \textit{\#fyp}, \textit{\#viral}, exhibits minimal internal structure with only 3.2\% of its edges occurring within the community, consistent with its function as a mechanism for algorithmic visibility rather than topical organization.

\begin{figure}[t]
    \centering
    \includegraphics[width=\linewidth]{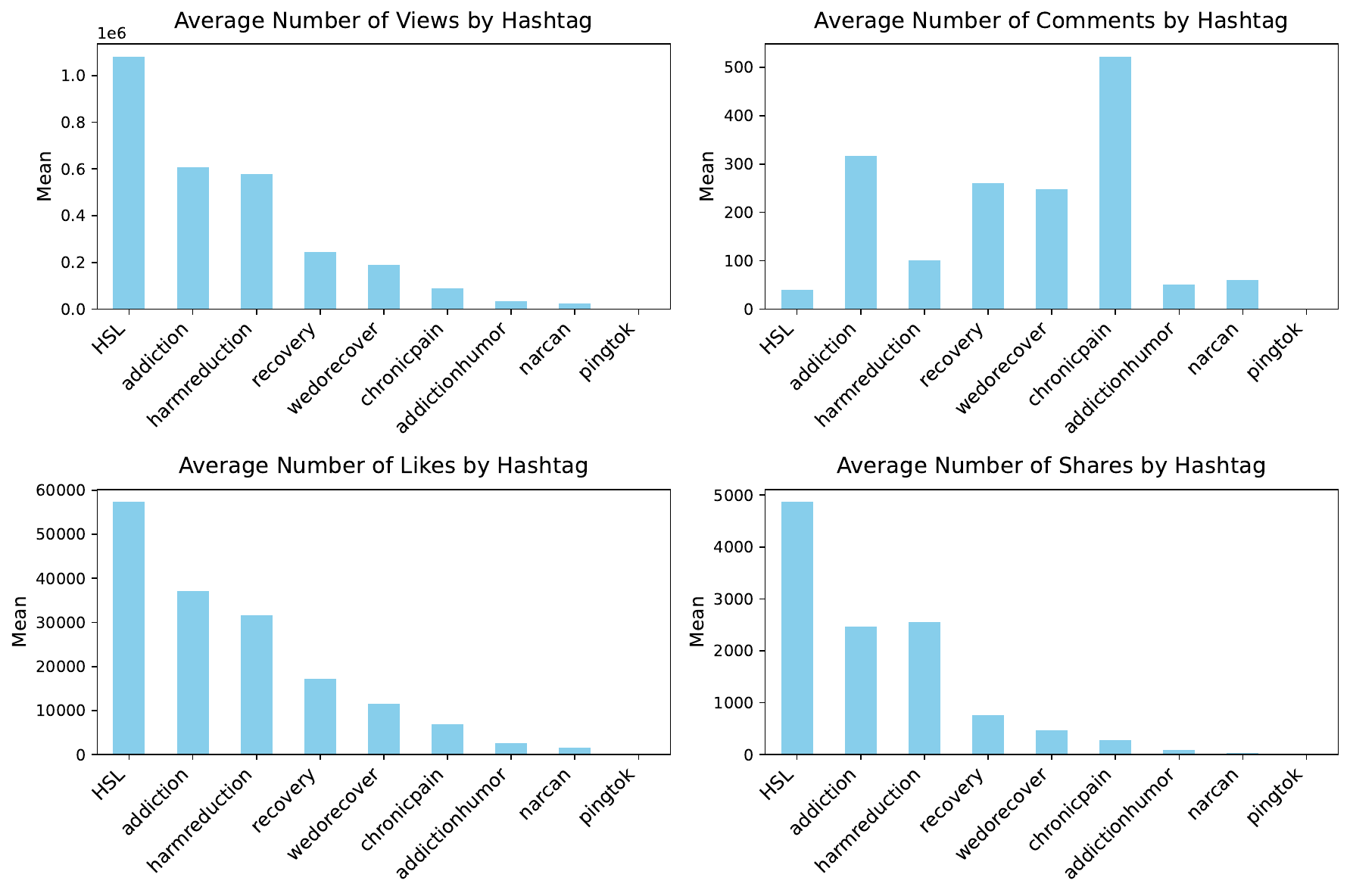}
    \caption{Engagement metrics across select popular hashtags in the coded set. Note, \#harmreductionsaveslives has been shorted to HSL.}
    \label{fig:engagement_hashtag}
\end{figure}

\subsection{RQ2: Video Topics and Substance Presence}

Analysis of 351 videos using the qualitative codebook (Table~\ref{tab:codebook}) revealed that the most prevalent category of content focused on advocacy and recovery support (33.9\%, $n=119$ videos). Videos in this category commonly featured creators sharing personal recovery journeys, celebrating sobriety milestones, or offering encouragement and support to others navigating substance use challenges. For example, one video documented a creator celebrating a ``100 days sober'' milestone while encouraging viewers to seek help.

Satirical and humorous content comprised the second largest category (28.2\%, $n=99$ videos). These videos frequently used comedy or parody, often through skits or reenactments (e.g., ``comedic skit about stereotypical behavior while intoxicated''), to engage with substance-related experiences. Informational content, often framed from claimed medical or professional perspectives (e.g., ``doctors explaining the effects of new medications on brain chemistry''), accounted for 13.1\% ($n=46$) of videos and included educational explanations of substance effects, risks, or treatment options. A smaller subset of videos focused primarily on platform engagement (3.9\%, $n=14$ videos), where substance-related themes were used mainly to attract views or followers. Promotional content appeared in 7.0\% ($n=24$) of videos, most commonly featuring demonstrations of vaping devices, Juul or lighter, as well as occasional promotion of nootropics or so-called study drugs.
Across all content categories, direct depiction of substances was comparatively limited. Substances were visually present in 25.9\% ($n=91$) of videos: 19.4\% ($n=68$) showed substances without active use, while only 6.6\% ($n=23$) depicted active substance use. This pattern suggests that substance-related discourse on TikTok more frequently involves discussion, narration, or symbolic reference rather than explicit visual display.

\begin{table}[t]
\small
\centering
\resizebox{\linewidth}{!}{
\begin{tabular}{|p{1.5cm}|p{2.3cm}|p{4.6cm}|p{2cm}|p{5.4cm}|}
\hline
\textbf{Dimension} & \textbf{Category} & \textbf{Definition} & \textbf{N (\%)} & \textbf{Example} \\ \hline

{Visible Substance} & Shown & Direct visual depiction of drugs or other substances &91(25.9\%),with 23(25\%) Actively Using & Displaying empty prescription medication bottles on a table beside the subject while they reminisce about prior SUD \\ \cline{2-5}
& Not Shown & No direct visual depiction of substances& 260 (74.1\%) & discussing recovery journey without showing any substances \\ \hline

 & Recovery Advocacy & Content promoting or raising awareness about recovery efforts & 119 (33.9\%) & Person sharing their 100 days sober milestone and encouraging others to seek help \\ \cline{2-5}
& Satirical & Humorous or satirical content about substance use without meaningful substance-related message & 99 (28.2\%) & Comedic skit about stereotypical behavior while intoxicated \\ \cline{2-5}
{Video Topic}& Informational & Educational content about substances from claimed professional/medical perspective & 46 (13.1\%) & Doctor explaining the effects of a new medication on brain chemistry \\ \cline{2-5}
& Promotional & Content meant to promote or sell substance-related products & 24 (6.8\%) & Video advertising nicotine vaping products with links to purchase \\ \cline{2-5}
& Social Documentation & Content created primarily for social media engagement, focusing on filming substance use & 14 (4.0\%) & Creator recording themselves taking shots at a party while encouraging viewers to like/follow \\ \cline{2-5}
& Trip Reports & First-person accounts of substance use experiences & 4 (1.1\%) & Individual describing their experience with a specific substance and its effects\\ \cline{2-5}
& Other & Substance-related content that doesn't fit other categories & 45 (12.8\%) & News clip about substance policy changes \\\hline

\end{tabular}
}
\caption{Codebook for qualitative analysis of substance-related TikTok videos (N=351)}
\label{tab:codebook}
\end{table}


We next examined how videos' content aligned with their associated hashtag and communities. While each video was coded for its primary topic, videos typically contained hashtags from multiple communities, revealing complex patterns in how substance-related content is tagged and disseminated. The following analysis focuses on the three most prevalent content categories: Advocacy and recovery, Entertainment, and Educational content - which together comprise 75.2\% of coded videos. Full cross-tabulations for all categories are provided in Table~\ref{tab:community_topic} in the Appendix.

\begin{figure}[t]
    \centering
    \includegraphics[width=0.7\linewidth]{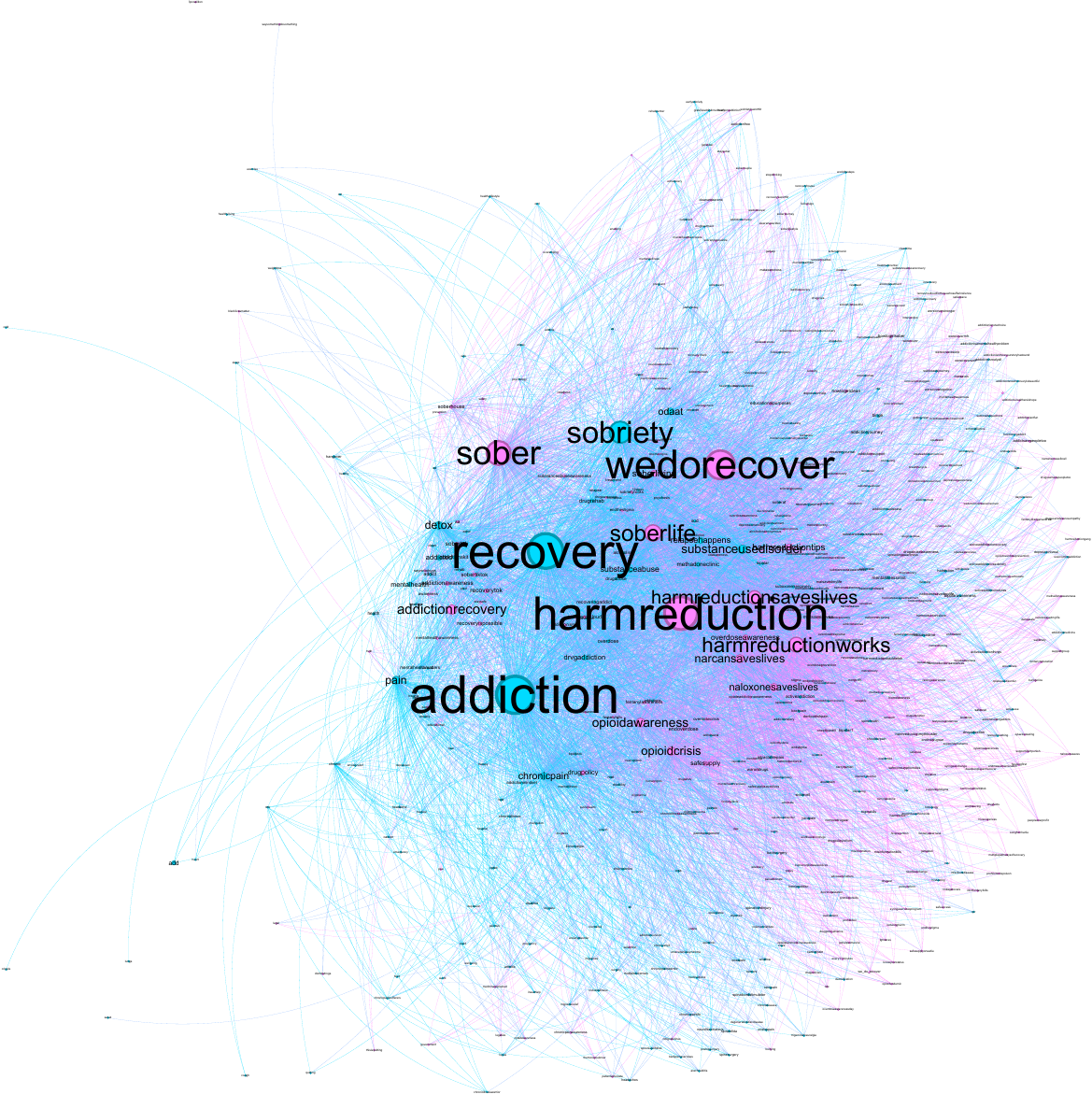}
    \caption{Sub-networks of Awareness \& Advocacy (Pink) and Health Conditions (Blue) to visualize strong overlap between hashtag usage.}
    \label{fig:awareness_health}
\end{figure}

\textbf{Advocacy, Recovery, and Support Content: } 
Videos coded primarily as advocacy and recovery content (33.9\%, $n=119$ videos), the largest thematic group in our manual coding, showed strong alignment with two major hashtag communities (see Table \ref{tab:community_topic} in Appendix):  

\begin{itemize}
   \item Awareness \& Advocacy (57.2\%, $n=99$ videos): Recovery-focused videos frequently used hashtags like \textit{\#harmreduction} and \textit{\#wedorecover}. Notably, while recovery-focused hashtags were more commonly used overall, content tagged with harm reduction messaging (particularly \textit{\#harmreductionsaveslives}) generated substantially higher user engagement across all metrics - views, likes, comments, and shares (Figure \ref{fig:engagement_hashtag}). 
    \item Health Conditions (56.8\%, $n=100$ videos): Videos discussing recovery often included medical or health-focused tags like \textit{\#addiction}, \textit{\#chronicpain}, and \textit{\#pain} (Figure \ref{fig:awareness_health})
\end{itemize}

Content in this group was typically centered on active or past efforts to recover from substance use, framed through cautionary narratives or expressions of care regarding the potential harms of substance misuse.

\textbf{Entertainment and Platform Engagement} Entertainment and humor-focused content (28.2\%, $n=99$ videos) as well as videos that documented using substances for engagement (4\%, $n=14$ videos) also bridged multiple communities. These videos predominantly contained hashtags from the following communities:

\begin{itemize}
    \item Platform (40.6\%, $n=82$ videos): Entertainment videos heavily utilized platform-specific tags like \textit{\#fyp} and \textit{\#pingtok}
    \item Humor (78.9\%, $n=60$ videos): Videos using \textit{\#addictionhumor} and similar tags often presented substance use through comedy or satire
    \item Substance Effects (53.8\%, $n=35$ videos): Videos that were satirical or relatable often documented an effect of substance use (e.g., what it's like being \textit{\#high} or \textit{\#stoned}). The subjects in these videos may not necessarily be intoxicated, but could be enacting or portraying the effects.
\end{itemize}
While substance use and misuse remained a topic in these videos, humor served as the dominant framing. A large portion of videos in this group were meme-based, in which non–drug-related content was reframed through captions, filters, or contextual cues, or drug-related content was presented primarily for comedic effect rather than instructional or cautionary purposes.

\begin{figure}
    \centering
    \includegraphics[width=0.7\linewidth]{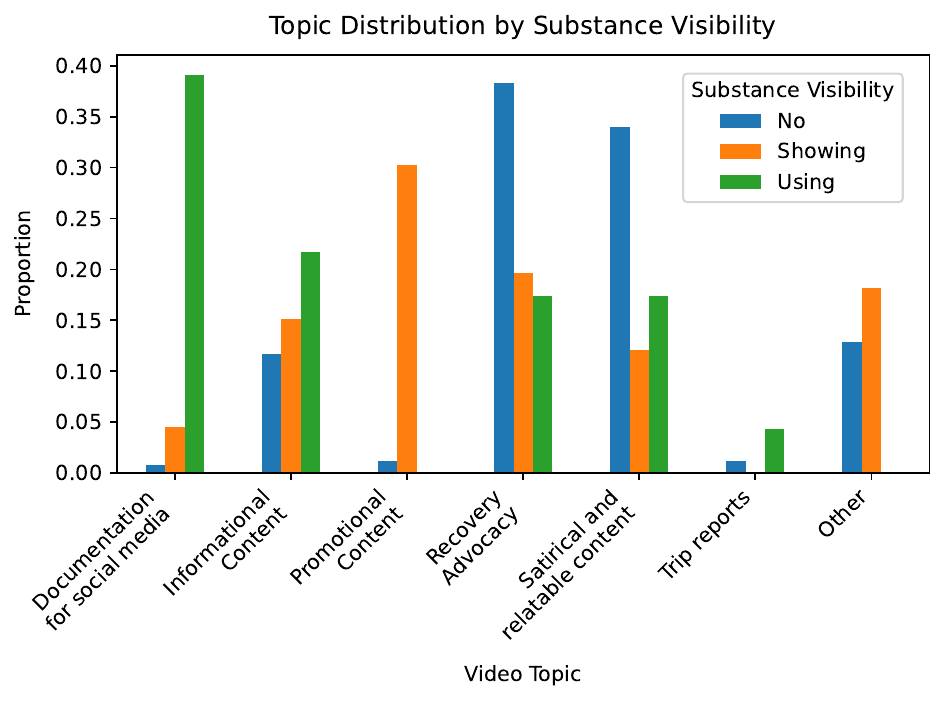}
    \caption{Distribution of video topics across three levels of substance visibility (no substances shown, substances shown but not used, and active substance use) in TikTok videos.}
    \label{fig:substance_presence}
\end{figure}

\textbf{Educational and Informational Content} Videos coded as informational (13.1\%, $n=46$ videos) showed connections to multiple professional and health-focused communities:
\begin{itemize}
    \item Health Conditions (11.4\%, $n=20$ videos): Medical information and health impacts
    \item Awareness \& Advocacy (16.8\%, $n=29$ videos): Harm reduction education
    \item Other Substances (24.6\%, $n=16$ videos) : 
    \item Occupation (19.2\%, $n=10$ videos): Content from healthcare workers and other professions
\end{itemize}

Content in this group was predominantly clinical or informational in tone. Humor was used sparingly, if at all. Most videos relied on detailed captions or longer-form explanations, often delivered by health professionals, recovery advocates, or organizations, many of which exhibited higher production value.

\textbf{Cross-Cutting Communities.}
While advocacy, entertainment, and educational content showed strong alignment with specific hashtag communities, certain clusters served as contextual bridges across all content types. Hashtags from the Commonly Misused Substances (31.6\%, $n=111$ videos), Identity \& Community (27.1 \%, $n=95$ videos), and Occupation (15.1 \%, $n=53$ videos) clusters appeared across all seven content categories, suggesting their role in providing context rather than signaling specific content types. For instance, videos tagged with \textit{\#opiates} spanned multiple content categories: healthcare workers explaining naloxone administration (Recovery/Advocacy), recovery nurses humorously discussing their reliance on caffeine (Satirical), and medical professionals explaining medication effects (Informational). Similarly, occupation-related hashtags like \textit{\#nurse} or \textit{\#doctor} and identity-based tags like \textit{\#disabled} or \textit{\#veteran} appeared across all content. Consumption Method hashtags like \textit{\#vape} or \textit{\#smoke} (28.8\%, $n=101$ videos) served a more specific bridging function, describing substance use behaviors across content types.

\subsection{RQ3: Effects of Substance Visibility}
Given the observed patterns in how content creators use hashtags to frame and contextualize their videos, we next examined how the visibility of substances or related paraphernalia within videos relates to content type and engagement patterns. Videos were categorized into three groups based on substance presence: no substances shown ($n=260$), substances shown but not actively used ($n=68$), and active substance use ($n=23$). Videos showing active substance use were predominantly focused on documentation for social media (39.1\% of such videos; Figure \ref{fig:substance_presence}), with a stronger tendency towards lifestyle or vlog-style content. In contrast, recovery advocacy content was most prevalent in videos without substance visibility (38.8\%), while promotional content appeared most frequently in videos that showed but did not depict active use of substances (29.4\%; e.g., videos promoting nicotine products). Within videos showing substances, content predominantly featured alcohol, nicotine, or cannabis products, with only one video collage containing ``hard drugs'' such as Percocet, Adderall, and Xanax. 

\emph{Kruskal-Wallis H} tests \cite{kruskal_use_1952} compared engagement metrics (views, likes, comments, and shares) across substance visibility categories (Table~\ref{tab:engagement_visibility}).  Results indicated a statistically significant difference in likes ($H=6.97$, $p=.031$), with videos depicting active substance use showing higher median likes (Mdn $= 13{,}000$) compared to videos showing substances without use (Mdn $= 982$) or no substances (Mdn $= 1{,}028$). Differences in views ($p=.087$), comments ($p=.171$), and shares ($p=.093$) did not reach statistical significance at $\alpha = .05$. However, descriptive patterns consistently showed higher median engagement for videos depicting active substance use across all metrics - notably, median views were over five times higher (79,100 vs. 14,900) and median comments nearly four times higher (150 vs. 39) compared to videos without substance visibility. Given the small sample size for active use videos ($n=23$), these results should be interpreted cautiously. Nevertheless, the findings suggest that the nature of substance depiction may influence audience engagement, with active use content attracting greater interaction despite comprising a small fraction of the overall substance-related content landscape.



\begin{table}[t]
\centering
\small
\begin{tabular}{lrrrrrr}
\toprule
 & \multicolumn{3}{c}{\textbf{Median (IQR)}} & & \\
\cmidrule(lr){2-4}
\textbf{Metric} & \textbf{No} & \textbf{Showing} & \textbf{Using} & \textbf{H} & \textbf{p} \\
 & $(n=260)$ & $(n=68)$ & $(n=23)$ & & \\
\midrule
Views & 14,900 & 22,150 & 79,100 & 4.88 & .087 \\
 & \footnotesize{(1,972--86,775)} & \footnotesize{(968--234,175)} & \footnotesize{(12,450--263,200)} & & \\[0.5em]
Likes & 1,028 & 982 & 13,000 & 6.97 & \textbf{.031*} \\
 & \footnotesize{(160--6,146)} & \footnotesize{(43--13,275)} & \footnotesize{(1,222--42,250)} & & \\[0.5em]
Comments & 39 & 34 & 150 & 3.53 & .171 \\
 & \footnotesize{(7--173)} & \footnotesize{(2--172)} & \footnotesize{(21--253)} & & \\[0.5em]
Shares & 20 & 24 & 112 & 4.76 & .093 \\
 & \footnotesize{(2--198)} & \footnotesize{(1--458)} & \footnotesize{(28--350)} & & \\
\bottomrule
\end{tabular}
\caption{Engagement metrics by substance visibility category. Kruskal-Wallis H tests compared engagement across visibility levels. IQR = Interquartile Range. *$p < .05$}
\label{tab:engagement_visibility}
\end{table}
\section{Discussion}
\label{sec:discussion}

Our findings present a critical case study of a socio-technical ecosystem where user agency and platform governance collide. While platforms like TikTok are often critiqued for amplifying harmful behaviors,  including eating disorders \cite{Bickham2024HiddenInPlainSight}, self-harm content \cite{lookingbill2024always}, and dangerous viral challenges \cite{curtis2018meta, rutherford_turnttrending_2023}, our analysis of its algorithmically surfaced content reveals a more complex reality. These insights emerge from a mixed-methods analysis that combines hashtag network structure with qualitative coding of algorithmically surfaced video content, allowing us to examine not only what content exists but what is made visible at scale. We find that, far from being a simple vector for promotion, the visible discourse on substance use is dominated by digital recovery narratives and harm reduction advocacy. This suggests that a combination of community-driven needs, user-led identity work, and active platform moderation has cultivated a powerful, large-scale recovery support space. This creates a fundamental tension: while the platform's recommendation algorithm has documented harms, this same algorithmic infrastructure enables the formation of robust recovery support communities. This ``algorithmic ambivalence'' \cite{phillips2018ambivalence} poses critical challenges for platform governance, as interventions designed to mitigate harm risk dismantling these beneficial support structures.

\subsection{The Politics of Visibility in Digital Recovery Narratives}
The dominance of recovery-focused content (33.9\% of videos) suggests that TikTok functions as a venue for what \citet{slade2021recorded} term ``digital recovery narratives.'' These narratives serve a dual function. For creators, the act of public declaration serves as a form of `stigma management' \cite{goffman2009stigma}, where the platform's affordances for visibility are used to reframe a marginalized identity into one of strength and advocacy. For viewers, these narratives offer realistic models of recovery success, offer avenues for connection \cite{Phelan2025SocialMediaRecovery}, and provide `social proof'  \cite{cialdini2004social} of sobriety, countering fatalistic views of addiction \cite{corvini_impact_2024}. This mirrors findings by \citet{li_explaining_2025}, who identified that ideological homogeneity and central agendas are key to sustained participation in online movements; here, the `movement' is recovery, and the shared identity is the driver of community.

However, this visibility raises questions regarding representation and power. While our findings demonstrate that recovery advocacy content achieves algorithmic prominence, understanding \textit{which} recovery narratives are amplified, and \textit{which} are marginalized, requires further investigation. The \citet{Phelan2025SocialMediaRecovery} finding that individuals in SUD recovery actively use platforms like TikTok for support underscores the stakes of these visibility dynamics \cite{Phelan2025SocialMediaRecovery}. The absence of active use content in our sample does not necessarily imply its absence from the platform, but rather its algorithmic suppression or relegation to private spaces \cite{gillespie2018custodians}. This reflects a form of ``uneven data shadows'' \cite{shelton2014mapping}, where certain users and narratives, potentially including those in active addiction, those from marginalized communities, or those whose experiences don't align with platform-favored content formats, remain algorithmically invisible. Consequently, claims about substance use on TikTok must be qualified as claims about what the platform's governance systems \textit{allow} to be widely visible.

\subsection{Humor as Boundary Work and Stigma Management}
The significant presence of humorous content (28.2\%) alongside serious recovery narratives reflects a distinct form of health communication that we conceptualize as ``boundary work'' \cite{gieryn1983boundary}. In this context, humor helps users navigate the boundaries of platform acceptability and social stigma. By framing substance use experiences through satire and memes  \cite{shifman2014memes}, creators may discuss addiction without the immediate vulnerability of serious or sensitive disclosure \cite{andalibi2017sensitive}, potentially lowering the barrier to entry for community participation. This finding aligns with \citeauthor{rifat_combating_2024}'s \cite{rifat_combating_2024} observation that community-specific cultural practices—in their case, religious traditions, and in ours, vernacular humor—are essential for mediating online conflict and harm.

While research has shown that comedy can help individuals process difficult experiences \cite{martin2002laughter}, the prevalence of humor in this domain can also introduce risks of minimization. The algorithmic favoring of high-engagement humorous content \cite{bucher2017algorithmic}  could pressure creators to frame serious health crises comedically to achieve visibility, potentially distorting the perceived severity of substance use disorders for outside observers \cite{chou2018addressing}. This tension between therapeutic humor and potential minimization represents a key challenge for platforms hosting health-related content, suggesting that engagement metrics alone are insufficient proxies for the quality of health support.

\subsection{Network Structure and the Cooperative Negotiation of Space}
Our network analysis challenges the prevailing narrative of ``adversarial algospeak,'' where users are assumed to be constantly evading moderation through coded language \cite{chancellor2016thyghgapp, klug2023algorithm}. While we observed some coded hashtags (e.g., \textit{\#dr00gtiktok}), the strong alignment between content types and straightforward hashtag communities (e.g., \textit{\#harmreduction}, \textit{\#wedorecover}) suggests that users are largely creating discourse within the bounds of platform affordances \cite{boyd2007social}, rather than purely against them. This indicates a cooperative, albeit fragile, negotiation of space where the community has successfully signaled the legitimacy of recovery discourse to the platform's moderation systems, whether through 
community signaling, policy evolution, or emergent algorithmic behavior.

Structurally, we found that recovery-focused hashtags function as essential ``bridge nodes''  \cite{burt1992structural} in the network, connecting otherwise siloed communities. This small-world \cite{watts_collective_1998} property enables efficient information flow between disparate groups, such as those focused on specific substances and those focused on sobriety. This structural insight is crucial for governance: blunt-force moderation strategies, such as banning specific keywords, could sever these bridges, fragmenting the network and isolating users in riskier sub-communities without pathways to support \cite{chandrasekharan2017you}.

\subsection{Implications for Harm-Aware Platform Design}
Our findings suggest that platform governance must move beyond a binary of ``remove'' or ``allow'' \cite{gillespie2018custodians} toward a strategy of context-sensitive amplification. We propose several concrete design directions:

\textbf{1- Leverage network topology for harm reduction pathways.} Our identification of recovery-focused hashtags as bridge nodes (\textit{\#addiction}, \textit{\#recovery}, \textit{\#harmreduction}) reveals structural leverage points for intervention. Rather than simply suppressing substance-related content, recommendation algorithms \cite{resnick1997recommender}  could proactively surface content from these bridge communities to users engaging with substance-specific or active-use clusters. For example, users repeatedly engaging with \textit{\#opiates} or \textit{\#pingtok} content could receive algorithmically-inserted suggestions for \textit{\#harmreduction} or \textit{\#wedorecover} content, creating pathways from substance-specific communities toward support resources.This approach aligns with public health principles of ``meeting people where they are''\cite{marlatt2012harm} while leveraging the small-world network structure \cite{watts_collective_1998}  we identified.

\textbf{2- Differentiate content types in moderation systems.} Our codebook (Table \ref{tab:codebook}) demonstrates that recovery advocacy, harm reduction information, active use depiction, and community humor are distinguishable categories.Current moderation systems often treat all substance-related content uniformly \cite{gillespie2018custodians}, but context-sensitive classification—potentially incorporating community input \cite{jhaver_does_2019} or subject-matter expertise—could enable more nuanced interventions. Content depicting active use without educational framing might be age-gated or de-amplified, while recovery narratives and harm reduction information could be preserved or even amplified. However, such systems risk creating new forms of surveillance \cite{eubanks2025automating} and must be designed with strong privacy protections and appeal mechanisms  \cite{klonick2018new}.

\textbf{3- Design for semi-private peer support.} The current ``public-by-default'' architecture of TikTok \cite{INPLP_TikTok_2024} creates tensions between visibility (for community-building) and privacy (to avoid stigma \cite{goffman2009stigma}). Platforms could offer middle-ground affordances: opt-in recovery communities visible to interested users but not broadcast to followers, pseudonymous accounts specifically for health disclosure, or fine-grained content controls allowing users to share recovery content with support networks while keeping it separate from professional/family audiences \cite{marwick2011tweet}. Such features would need careful design to avoid inadvertently marking recovery content as shameful.

\textbf{4- Involve affected communities in governance.} Our finding that hashtag communities show organic alignment rather than adversarial evasion suggests potential for collaborative governance \cite{suzor2019we}. Rather than purely top-down moderation, platforms could establish advisory structures including people with lived experience, harm reduction advocates, and public health professionals. Such approaches could help navigate difficult trade-offs between protecting users and preserving support communities, while ensuring governance decisions don't reproduce existing power inequities in who gets to define ``appropriate'' recovery discourse \cite{noble2018algorithms}.

Critical to all these recommendations is recognition that the same affordances enabling harm also enable help. Well-intentioned changes risk unintended consequences \cite{friedman2013value}: special treatment could stigmatize recovery content, surveillance infrastructures built for safety could be weaponized against marginalized users \cite{eubanks2025automating}, and clinical norms imposed on peer communities could undermine their organic character \cite{solomon2004peer}. Effective governance requires explicitly designing for this duality rather than seeking one-sided solutions.

\subsection{The Politics of Algorithmic Invisibility}

Our methodology, analyzing algorithmically-served, publicly visible content, captures what TikTok's governance systems allow to surface \cite{gillespie2018custodians}, not the totality of substance-related discourse. This is not merely a limitation but a finding: the dominance of recovery content in our sample reflects active platform curation \cite{gillespie2018custodians} through moderation policies and algorithmic amplification. 

What remains invisible reveals as much about platform governance as what is visible \cite{shelton2014mapping}. Content in private accounts, shadow-banned videos \cite{gillespie2010politics}, age-gated material, and algorithmically suppressed posts likely contain different narratives—potentially including active use content, harm reduction information deemed ``controversial,'' or experiences from marginalized users whose content may not align with algorithmically-favored 
formats or whose access to platform visibility is constrained. The absence of active use depiction in our sample (only 6.5\%) does not prove TikTok is free of such content \cite{bishop2018anxiety}; rather, it demonstrates that TikTok's recommendation system successfully prevents this content from wide circulation.

This raises critical questions: Should platforms amplify certain recovery modalities over harm reduction approaches? Who decides what constitutes ``appropriate'' substance use discourse? Our data suggests TikTok has already made these decisions, whether through explicit policy or emergent algorithmic behavior \cite{gillespie2010politics}, in ways that privilege certain recovery modalities and demographic groups. The most vulnerable users—those in active addiction, those with intersecting marginalized identities \cite{crenshaw1991mapping}, those in acute crisis—may be least likely to appear in algorithmically-surfaced public content \cite{noble2018algorithms}, raising concerns about whose needs platform governance serves.

\section{Limitations}
Our findings must be interpreted within several important constraints. First, our findings may not fully represent the entire TikTok ecosystem, as they are influenced by limitations in data availability and collection processes. TikTok's (unofficial) API serves content through its recommendation algorithm, which means we received videos that did not necessarily contain our queried hashtags. While we attempted to address this by sampling videos across engagement quartiles to ensure representativeness, the algorithm's opacity makes it difficult to guarantee complete coverage of substance-related content. This algorithmic bias may have also led to over-representation of certain creators whose content is favored by TikTok's recommendation system. TikTok, in our experience, also cuts off data requests arbitrarily. Our approach of using 1,000 videos per hashtag was due to an alignment with the average number of videos we could reliably retrieve before being cut off on that request. Second, the dynamic nature of social media data presented challenges - some videos included in our initial network analysis were later unavailable for content coding due to accounts being banned or made private, a particular issue in communities discussing sensitive topics like substance use. Third, methodological challenges arose around identifying substance-related content, as users often employ common words as coded references to substances, making it difficult to definitively categorize some content as substance-related versus general discussion. This challenge was compounded by the platform's rapid evolution of slang and terminology. Additionally, some videos contained duplicate hashtags, which could affect network metrics, and less popular hashtags may be underrepresented in our sample despite efforts to ensure broad coverage. Lastly, our study analyzed content characteristics rather than user impacts. We cannot make claims about whether exposure to recovery content affects help-seeking behaviors, treatment outcomes, or relapse risks. The relationship between what people see on TikTok and their subsequent actions remains an open question. These limitations underscore a fundamental point: platform research always captures a mediated reality shaped by technical systems, corporate governance, and social dynamics. Claims about ``what happens on TikTok'' must be qualified as ``what TikTok's governance systems allow to be widely visible.''

\section{Conclusion and Future Direction}
\label{sec:conclusion}

This study challenges narratives about social media platforms and substance use. Through analysis of 2,333 hashtags and 351 videos, we found that TikTok's algorithmically surfaced, publicly visible substance use content is dominated by recovery advocacy (33.9\%) and humor-based coping (28.2\%), not glamorization or promotion. Our network analysis revealed a highly interconnected small-world structure (clustering coefficient 0.52, average path length 2.20) where recovery-focused hashtags like \textit{\#addiction}, \textit{\#recovery}, and \textit{\#sober} function as critical bridges between thematic communities. Through manual coding of 351 representative videos, we found that Recovery Advocacy content dominated (33.9\% of videos), followed by Satirical content (28.2\%), while direct substance depiction appeared in only 26\% of videos, with active use shown in just 6.5\%—suggesting that despite documented harms from algorithmic recommendation systems, these same systems can enable valuable peer support communities around stigmatized health conditions.

However, our findings also reveal critical tensions inherent in platform governance. The dominance of recovery content in our sample reflects TikTok's active curation through moderation policies and algorithmic amplification—raising questions about whose narratives are privileged and what remains invisible. The alignment between hashtag communities and content types indicates organic community formation rather than adversarial evasion, suggesting users have successfully negotiated space within platform affordances. Yet this cooperative relationship remains fragile: changes in moderation policy, algorithmic updates, or increased regulatory pressure could easily disrupt these communities, pushing discourse underground or fragmenting support networks.

Our findings open several critical avenues for future research. While we found evidence of TikTok serving as a positive space for recovery support, there is a pressing need to understand the real-world impacts of this online discourse. Future studies should examine how exposure to recovery content affects help-seeking behaviors and treatment outcomes. Additionally, researchers should investigate potential risks and benefits for people who use drugs who engage with or create TikTok content. This is particularly important for content creators who publicly identify as being in active recovery, as sharing personal narratives may have complex psychological and social implications that are not yet well understood \cite{rennick-egglestoneMentalHealthRecovery2019}. Additionally, investigating the intersection of online and offline support networks could provide insights into how digital recovery communities complement traditional recovery support services, optimizing online and in-person interventions. This work would be particularly valuable for reaching younger populations, who may be more likely to seek initial support through social media platforms. 

\section{Ethics Statement}
This research was deemed exempt by our institution's IRB as it analyzes publicly available data with no interaction with human subjects. In our analysis of TikTok videos, we followed established practices for ethical social media research \cite{fiesler2020no} and we took several steps to protect user privacy: (1) all analysis was conducted on aggregated data, (2) no usernames or personally identifiable information are included in our results, and (3) our public dataset will only contain video URLs and associated metadata, allowing other researchers to hydrate the data while respecting user privacy. All collected and annotated data was stored on secure servers approved by our institution. These steps ensure that potential negative outcomes due to the use of these data are minimized while maintaining the reproducibility of our research. The dataset and associated code will be made available upon acceptance.

\bibliographystyle{ACM-Reference-Format}
\bibliography{aaai25, references_}

\newpage
\appendix
\setcounter{table}{0}
\setcounter{figure}{0}
\setcounter{section}{0}
\renewcommand{\thetable}{A.\arabic{table}}
\renewcommand{\thesection}{A.\arabic{section}}
\renewcommand{\thefigure}{A.\arabic{figure}}
\section{Full Hashtag Network with Communities}
\label{sec:network}

\begin{figure*}[ht]
\centering
    \includegraphics[width=\textwidth]{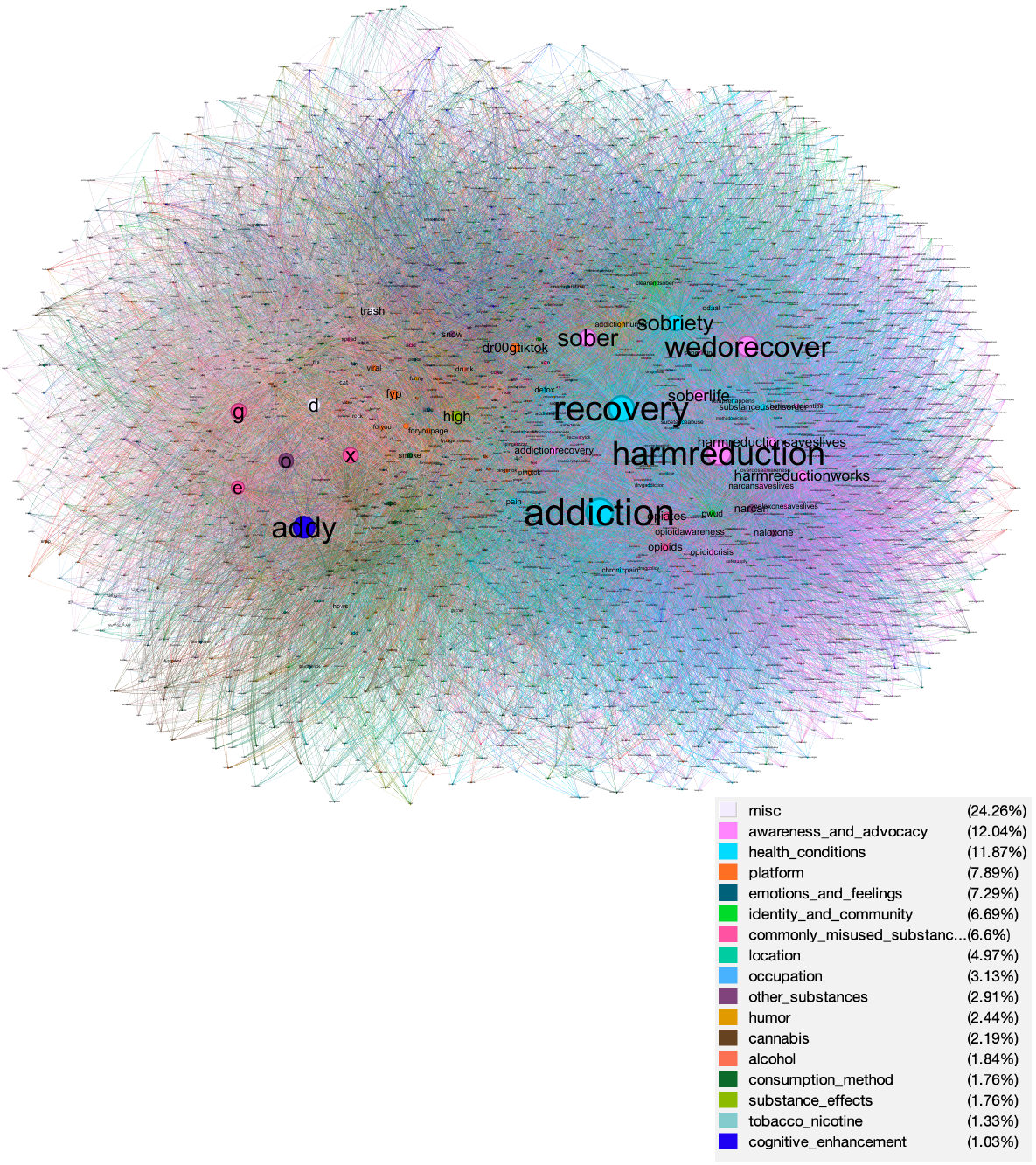}
    \caption{The overall network of 2,333 hashtags related to substance use with colors representing each community as shown in the legend}
    \label{fig:full-graph}

\end{figure*}

Figure \ref{fig:full-graph} shows a visualization of the complete hashtag co-occurrence network, containing 2,333 nodes (hashtags) and 46,990 edges. The nodes are colored according to their community membership as determined by our semi-automated community detection process. The network layout was generated using a Fruchterman-Reingold algorithm, which positions nodes with stronger connections closer together. The central position of recovery-focused hashtags like \textit{\#addiction}, \textit{\#recovery}, and \textit{\#harmreduction} is clearly visible, demonstrating their role as bridges between communities. Node sizes are scaled according to their degree centrality, highlighting the most connected hashtags in the network. This visualization helps illustrate both the small-world properties of the network (clustering coefficient 0.52, average path length 2.20) and the organic formation of thematic communities around substance-related discourse on TikTok.

\section{Network Centrality Measures per Community}

Table \ref{tab:centrality-measures} presents detailed network statistics and centrality measures for both the complete network and each of the 17 identified communities including Misc. For each community, we report basic network metrics (number of nodes, edges, average degree, weighted degree, clustering coefficient, and average shortest path length) as well as the top 5 most central nodes according to four different centrality measures (betweenness, closeness, degree, and eigenvector centrality). The whole network shows high connectivity (average degree 40.28) and clustering (0.52), with core hashtags like ``addiction", ``recovery", and ``sober" consistently ranking highly across all centrality measures. 

\begin{table*}[ht]
\centering
\footnotesize
\resizebox{\textwidth}{!}{%
\begin{tabular}{@{}p{2.2cm}rrrrrrrp{2.2cm}p{2.2cm}p{2.2cm}p{2.2cm}@{}}
\toprule
\textbf{Network} &
  \textbf{\#N} &
  \textbf{\#E} &
  \textbf{Avg D} &
  \textbf{Avg WD} &
  \textbf{Avg C} &
  \textbf{Avg SP} &
  \textbf{Avg B} &
  \textbf{Betweenness} &
  \textbf{Closeness} &
  \textbf{Degree} &
  \textbf{Eigenvector} \\ \midrule
Whole Network &
  2333 &
  46990 &
  40.28 &
  513.84 &
  0.52 &
  2.20 &
  0.0005 &
  addiction, recovery, sober, wedorecover, harmreduction &
  addiction, recovery, sober, wedorecover, harmreduction &
  addiction, recovery, sober, wedorecover, harmreduction &
  addiction, recovery, sober, wedorecover, harmreduction \\ \midrule
Awareness \& Advocacy &
  279 &
  2164 &
  15.51 &
  334.30 &
  0.89 &
  1.98 &
  0.0010 &
  harmreduction, wedorecover, harmreductionsaveslives, harmreductionworks, sober &
  harmreduction, harmreductionsaveslives, wedorecover, harmreductionworks, sober &
  harmreduction, harmreductionsaveslives, wedorecover, harmreductionworks, sober &
  harmreduction, harmreductionsaveslives, wedorecover, harmreductionworks, narcansaveslives \\ \midrule
Health Conditions &
  277 &
  1522 &
  10.99 &
  224.09 &
  0.85 &
  2.02 &
  0.0012 &
  addiction, recovery, sobriety, pain, chronicpain &
  addiction, recovery, sobriety, pain, chronicpain &
  addiction, recovery, sobriety, pain, chronicpain &
  addiction, recovery, sobriety, pain, detox \\ \midrule
Emotions \& Feelings &
  172 &
  305 &
  3.55 &
  20.65 &
  0.31 &
  2.39 &
  0.0002 &
  peace, onedayatatime, friend, dream, serenity &
  peace, onedayatatime, dream, serenity, friend &
  peace, onedayatatime, dream, friend, serenity &
  peace, onedayatatime, dream, serenity, friend \\ \midrule
Platform &
  183 &
  146 &
  1.60 &
  39.58 &
  0.12 &
  2.44 &
  0.0004 &
  pingtok, dr00gtiktok, pingertok, chat, fyp &
  pingtok, dr00gtiktok, pingertok, fyp, fyp (shi) &
  pingtok, dr00gtiktok, pingertok, chat, stonertokfyp &
  pingtok, dr00gtiktok, pingertok, chat, fyp \\ \midrule
Identity \& Community &
  155 &
  97 &
  1.25 &
  8.22 &
  0.09 &
  3.21 &
  0.0001 &
  cleanandsober, na, pwud, 30s, women &
  pwud, na, cleanandsober, women, lgbt &
  cleanandsober, pwud, na, 30s, mujeres &
  cleanandsober, pwud, na, lgbtqia, calisober \\ \midrule
Commonly Misused Substances &
  152 &
  553 &
  7.28 &
  60.47 &
  0.38 &
  2.47 &
  0.0007 &
  opioids, opiates, coke, molly, fent &
  opioids, coke, opiates, molly, uppers &
  opioids, opiates, coke, molly, fent &
  opioids, coke, opiates, fent, molly \\ \midrule
Location &
  116 &
  19 &
  0.33 &
  1.26 &
  0.00 &
  1.80 &
  <0.0001 &
  chinatown, paradise, yucatan, arnolds, filmteyvatislands &
  chinatown, paradise, india, china, texas &
  chinatown, paradise, yucatan, arnolds, india &
  chinatown, india, china, texas, egypt \\ \midrule
Cannabis &
  51 &
  229 &
  8.98 &
  39.25 &
  0.55 &
  2.04 &
  0.0006 &
  maryjane, weed, cannabis, skunk, herb &
  maryjane, weed, cannabis, herb, grass &
  maryjane, weed, cannabis, herb, grass &
  maryjane, cannabis, weed, herb, grass \\ \midrule
Alcohol &
  43 &
  92 &
  4.28 &
  56.19 &
  0.41 &
  1.79 &
  0.0006 &
  drunk, tipsy, hammered, shitfaced, bars &
  drunk, hammered, tipsy, shitfaced, bars &
  drunk, hammered, tipsy, shitfaced, bars &
  drunk, hammered, tipsy, bars, shitfaced \\ \midrule
Cognitive Enhancement &
  24 &
  38 &
  3.17 &
  45.00 &
  0.46 &
  1.64 &
  0.0001 &
  nootropics, modafinil, smartdrugs, smartpills, focus &
  nootropics, smartdrugs, modafinil, smartpills, focus &
  nootropics, smartdrugs, modafinil, smartpills, focus &
  nootropics, smartdrugs, modafinil, smartpills, focus \\ \midrule
Substance Effects &
  41 &
  33 &
  1.61 &
  23.76 &
  0.16 &
  2.42 &
  0.0005 &
  high, wasted, energy, blackout, clarity &
  high, wasted, lit, crash, shadowpeople &
  high, wasted, blackout, lit, downers &
  high, wasted, lit, downers, crash \\ \midrule
Consumption Method &
  41 &
  62 &
  3.02 &
  89.80 &
  0.30 &
  1.86 &
  0.0017 &
  smoke, vape, smoking, vaper, vaping &
  smoke, vaper, vape, vaping, vaped &
  smoke, vaper, vape, vaping, vaped &
  smoke, vaper, vaping, vaped, vape \\ \midrule
Tobacco Nicotine &
  31 &
  115 &
  7.42 &
  68.90 &
  0.70 &
  1.70 &
  0.0004 &
  ecigs, ecig, cigarette, nicotine, juul &
  ecigs, cigarette, ecig, nicotine, juul &
  ecigs, cigarette, ecig, nicotine, juul &
  cigarette, ecig, ecigs, nicotine, juul \\ \midrule
Other Substances &
  70 &
  130 &
  3.71 &
  50.11 &
  0.27 &
  2.69 &
  0.0012 &
  narcan, snow, psychedelictok, naloxone, naltrexone &
  narcan, psychedelictok, snow, ��, shroooms &
  psychedelictok, snow, narcan, mushrooms, naloxone &
  psychedelictok, mushrooms, salvia, mescaline, magicmushroomsadventures \\ \midrule
Humor &
  56 &
  19 &
  0.68 &
  28.64 &
  0.00 &
  1.90 &
  0.0002 &
  addictionhumor, prank, pingtok⚫️��⚫️, standupcomedy, satire &
  addictionhumor, jokes, comedia, soberhumor, comediahumor &
  addictionhumor, jokes, comedia, soberhumor, comediahumor &
  addictionhumor, jokes, comedia, soberhumor, comediahumor \\ \midrule
Occupation &
  73 &
  5 &
  0.14 &
   0.93 &
  0.00 &
  1.50 &
  <0.0001 &
  copilots, truckdrivers, actress, nursing, outreachworker &
  copilots, truckdrivers, aviation, construction, creator &
  copilots, truckdrivers, aviation, work, construction &
  copilots, aviation, construction, creator, truckdrivers \\ \midrule
Misc &
  565 &
  2367 &
  8.38 &
  139.01 &
  0.29 &
  2.46 &
  0.0002 &
  cat, fire, rock, g, pink &
  cat, fire, g, rock, pink &
  cat, fire, g, rock, x &
  cat, fire, g, x, rock \\ \bottomrule
\end{tabular}}
\caption{Centrality measures for the whole network and each community. \#N=Number of Nodes, \#E=Number of Edges, Avg D=Average Degree, Avg WD=Average Weighted Degree, Avg C=Average Clustering, Avg SP=Average Shortest Path, Avg B=Average Betweenness Centrality. The top five nodes are presented per centrality measure. Consumption Method exhibits the highest average betweenness (bolded).}
\label{tab:centrality-measures}
\end{table*}

\section{Hashtag Community Distribution per Content Topic}
Table \ref{tab:community_topic} presents a cross-tabulation of hashtag communities and video content topics, allowing us to examine how different communities align with various types of content. This analysis reveals notable patterns in how hashtags from different communities are employed across content types. Recovery Advocacy content, the most prevalent in our dataset, shows strong representation across multiple hashtag communities, particularly drawing from the Awareness \& Advocacy and Health Conditions communities. Satirical content, our second most common category, shows interesting patterns of hashtag usage, frequently incorporating hashtags from Platform and Humor communities while also drawing from more serious communities - suggesting complex layering of messaging in humorous content. Informational content shows strong alignment with Professional and Health Condition hashtags, while Promotional content primarily uses hashtags from Consumption Method and Tobacco/Nicotine communities. This distribution helps validate our community detection approach by demonstrating meaningful alignment between hashtag communities and content types, while also revealing how creators strategically combine hashtags from multiple communities to reach diverse audiences.

\begin{table*}[ht]
\centering
\resizebox{\textwidth}{!}{%
\begin{tabular}{@{}l|cccccccc@{}}
\toprule
\textbf{Communities} &
  \textbf{\begin{tabular}[c]{@{}c@{}}Documentation of \\Use for \\Social Media\end{tabular}} &
  \textbf{Informational Content} &
  \textbf{Other} &
  \textbf{Promotional Content} &
  \textbf{Recovery Advocacy} &
  \textbf{\begin{tabular}[c]{@{}c@{}}Satirical and \\Relatable Content\end{tabular}} &
  \textbf{Trip Reports} &
  \textbf{Total} \\ \midrule
\textbf{Platform}                     & 8 (4.0\%)   & 20 (9.9\%)   & 25 (12.4\%) & 15 (7.4\%)  & 59 (29.2\%)  & 74 (36.6\%)  & 1 (0.5\%) & 202 \\
\textbf{Health Conditions}            & 0 (0.0\%)   & 20 (11.4\%)  & 17 (9.7\%)  & 3 (1.7\%)   & 100 (56.8\%) & 34 (19.3\%)  & 2 (1.1\%) & 176 \\
\textbf{Awareness and Advocacy}       & 1 (0.6\%)   & 29 (16.8\%)  & 12 (6.9\%)  & 1 (0.6\%)   & 99 (57.2\%)  & 29 (16.8\%)  & 2 (1.2\%) & 173 \\
\textbf{Misc}                         & 4 (3.2\%)   & 8 (6.4\%)    & 24 (19.2\%) & 7 (5.6\%)   & 42 (33.6\%)  & 39 (31.2\%)  & 1 (0.8\%) & 125 \\
\textbf{Commonly Misused Substances}  & 1 (0.9\%)   & 13 (11.8\%)  & 20 (18.2\%) & 1 (0.9\%)   & 39 (35.5\%)  & 36 (32.7\%)  & 0 (0.0\%) & 110 \\
\textbf{Consumption Method}           & 3 (3.0\%)   & 11 (11.1\%)  & 12 (12.1\%) & 14 (14.1\%) & 22 (22.2\%)  & 37 (37.4\%)  & 0 (0.0\%) & 99  \\
\textbf{Emotions and Feelings}        & 4 (4.1\%)   & 8 (8.2\%)    & 14 (14.3\%) & 4 (4.1\%)   & 32 (32.7\%)  & 36 (36.7\%)  & 0 (0.0\%) & 98  \\
\textbf{Identity and Community}       & 2 (2.1\%)   & 8 (8.5\%)    & 9 (9.6\%)   & 2 (2.1\%)   & 51 (54.3\%)  & 20 (21.3\%)  & 2 (2.1\%) & 94  \\
\textbf{Humor}                        & 0 (0.0\%)   & 2 (2.6\%)    & 4 (5.3\%)   & 1 (1.3\%)   & 8 (10.5\%)   & 60 (78.9\%)  & 1 (1.3\%) & 76  \\
\textbf{Other Substances}             & 2 (3.1\%)   & 16 (24.6\%)  & 11 (16.9\%) & 0 (0.0\%)   & 21 (32.3\%)  & 15 (23.1\%)  & 0 (0.0\%) & 65  \\
\textbf{Substance Effects}            & 7 (10.8\%)  & 4 (6.2\%)    & 13 (20.0\%) & 4 (6.2\%)   & 6 (9.2\%)    & 28 (43.1\%)  & 3 (4.6\%) & 65  \\
\textbf{Location}                     & 3 (5.0\%)   & 4 (6.7\%)    & 13 (21.7\%) & 2 (3.3\%)   & 9 (15.0\%)   & 28 (46.7\%)  & 1 (1.7\%) & 60  \\
\textbf{Alcohol}                      & 8 (13.6\%)  & 6 (10.2\%)   & 6 (10.2\%)  & 0 (0.0\%)   & 19 (32.2\%)  & 18 (30.5\%)  & 2 (3.4\%) & 59  \\
\textbf{Tobacco/Nicotine}             & 2 (3.4\%)   & 9 (15.3\%)   & 5 (8.5\%)   & 12 (20.3\%) & 16 (27.1\%)  & 15 (25.4\%)  & 0 (0.0\%) & 59  \\
\textbf{Occupation}                   & 1 (1.9\%)   & 10 (19.2\%)  & 9 (17.3\%)  & 1 (1.9\%)   & 22 (42.3\%)  & 9 (17.3\%)   & 0 (0.0\%) & 52  \\
\textbf{Cannabis}                     & 0 (0.0\%)   & 3 (8.1\%)    & 8 (21.6\%)  & 0 (0.0\%)   & 4 (10.8\%)   & 21 (56.8\%)  & 1 (2.7\%) & 37  \\
\textbf{Cognitive Enhancement}        & 1 (4.0\%)   & 7 (28.0\%)   & 3 (12.0\%)  & 7 (28.0\%)  & 4 (16.0\%)   & 3 (12.0\%)   & 0 (0.0\%) & 25  \\ \bottomrule
\end{tabular}%
}
\caption{Distribution of communities corresponding with each hashtag used in the 351 analyzed videos per content topic. }
\label{tab:community_topic}
\end{table*}

\end{document}